\documentclass{article}
\usepackage[preprint]{neurips_2026}

\usepackage[utf8]{inputenc} 
\usepackage[T1]{fontenc}    
\usepackage{hyperref}       
\usepackage{url}            
\usepackage{booktabs}       
\usepackage{amsfonts}       
\usepackage{nicefrac}       
\usepackage{xcolor}         
\usepackage{amsmath}
\usepackage{amsthm}
\usepackage{graphicx}
\usepackage{amssymb}
\usepackage{algorithm}
\usepackage{algpseudocode}
\usepackage{float}
\usepackage{enumitem}
\usepackage{multirow}
\usepackage{wrapfig}
\usepackage{capt-of}
\usepackage{tabularx}
\usepackage{array}

\DeclareFontShape{T1}{ptm}{m}{scit}{<->ssub*ptm/m/sc}{}{}

\theoremstyle{remark}

\title{World State Generator}

\author{
    \begin{tabular}{ccc}
        \normalfont\mdseries SUNGHEON JEONG$^1$ &
        \normalfont\mdseries SANGGEON YUN$^1$
        \normalfont\mdseries RYOZO MASUKAWA$^{1}$ &
    \end{tabular}
    \\[0.7em]
    \begin{tabular}{cc}
        \normalfont\mdseries HALEH ALIMOHAMADI$^1$
        \normalfont\mdseries MAHDI IMANI$^2$ &
        \normalfont\mdseries MOHSEN IMANI$^1$
    \end{tabular}
    \\[1.0em]
    \begin{tabular}{c}
        $^{1}$University of California, Irvine, $^{2}$Northeastern University \\
        {\tt\small sungheoj@uci.edu}
    \end{tabular}
}

\begin{document}
    \maketitle

    \begin{abstract}
    Language agents solve complex tasks through plans and actions. A single step the world refuses puts the goal out of reach, and what the agent does next decides the task. Prompted planners fail at exactly this point, rewriting the refused step in new words, meeting the same refusal, and burning the attempt budget without moving. They fail because the plan was never tied to the world, so a refusal has nothing in the plan to attach to. A world is where a task runs, and it has its own rules, its own admissible actions, and its own constraints. We build synthetic worlds across 7 domains and extract training data from them. A program enforces each world's rules and grades its goal, and every world is admitted only if its goal is reachable from its initial state. Agents run inside and leave verified failures paired with repairs that carried the run to a state the world certified, a record of about 226K trajectories. On this record we train the \textit{World State Generator}, a model that writes a plan as checkable states of the world and keeps that plan aligned with the world it runs in. That alignment is what a plan written in language lacks, since the world it runs in has physical limits, logical dependencies, and required orders the language never states, and the plan encounters these rules only when a state fails. \textit{WSG} takes that failure as the rule the world has stated and rewrites the remaining states to obey it, so the plan bends to the world as the run goes on. Across 7 public benchmarks, \textit{WSG} raises end-to-end success for two open models near 30B parameters over prompting and brings to the level of proprietary model.
\end{abstract}
    \section{Introduction}
    A language agent working through a long task follows a plan, and long tasks give the world many chances to refuse a step of it. When that happens, the agent rewrites the refused step in different words, submits it again, and meets the same refusal, cycling until the attempt budget is spent \citep{huang2024large,yan2026tide}. Reflecting on the failure before the rewrite does not break the cycle \citep{shinn2023reflexion,madaan2023self}, which survives because the plan was never tied to the world. Since a prompted planner writes what it imagines the task should look like, nothing in that plan says which tools exist, which calls the interface accepts, or which conditions must already hold. The model therefore has nothing in its plan to attach the world's report to, and when that report comes, the only thing left to change is the wording. A refusal is therefore information, the world stating a rule the plan omitted, and repair means reading that rule and rewriting the rest of the plan to obey it.

    Learning to repair this way requires data with three properties. Every failure must come with the repair that resolved it, and both sides must be decided against the world rather than by opinion. Agent corpora are filtered for success, so failures are rarely kept at all \citep{zeng2024agenttuning,chen2023fireact}, generated environments that collect trajectories at scale harvest successes in the same way \citep{hu2025agentgen,song2026envscaler,guo2025genenv}, and a failure that does survive stands alone. Even a failure and a repair recorded together would not be decided, since no procedure checks whether a written state holds in its world, so nothing establishes that the failure was real or that the repair worked. The third property concerns where the pairs come from. A model trained on one world learns that world's rules, and rules are what differs between worlds, so the pairs must span many worlds for the model to learn how to read a refusal instead. Agents trained in one environment degrade sharply when the evaluation environment changes \citep{li2026benchmark}.

    No existing corpus holds pairs with these three properties, so we build them ourselves. We construct synthetic worlds, self-contained environments whose rules are enforced by program, and run agents inside them under a runtime that checks each state against the world. Two choices follow from what classical planning already knows. A world is written in a form a program can execute, so a solver can confirm that its goal is reachable from its initial state before any agent runs in it \citep{liu2023llm+,guan2023leveraging}, and each world is composed from seed material rather than stamped from one template \citep{hu2025agentgen}. The worlds are drawn from 7 domains, whose states record what must be known, what must be built, or how the world must stand, and no two worlds in a domain enforce the same rules. Agents running inside fail against the rules and repair their plans, and we keep a repair only when it carried the run to a state the world certified. Each kept pair holds a failure and the repair that followed it, both decided by the world rather than by us, and the pairs come from worlds that differ across all 7 domains.

    On these pairs we train the World State Generator (\textit{WSG}), a planner that keeps its plan aligned with the world it runs in. Alignment is what a prompted planner cannot hold, since the rules that decide a plan are physical limits, logical dependencies, and required orders that the goal never states, and the planner meets them one at a time as its steps fail. The runtime that collected these pairs is the State-Centric Decision Process (\textit{SDP})~\citep{jeong2026state}, where each step of a plan is written as a state the world must reach, so the loop checks each state against the world and reports which condition did not hold and what it found instead, a refutation. \textit{SDP} fills the planner in this loop by prompting, and \textit{WSG} is a model trained to fill it, so the loop is unchanged and only the planner inside it learns. We train it on the repair call alone, so the first plan of a run is still written by the prompted backbone. \textit{WSG} reads that refutation as a rule of the world and rewrites the remaining states to obey it, so the plan bends to the world as the run goes on. Given a goal, the states the world has already certified, and the refutation, \textit{WSG} returns the states that carry the world from where it stands to the goal.

    Our contribution is the pipeline that produces these pairs at scale, the model trained on them, and the finding that repair learned in synthetic worlds transfers to worlds outside the corpus. We construct 226K trajectories in this way and train \textit{WSG} on them in open models near 30B parameters. The evaluation is zero-shot, since none of the benchmarks contributed a world, a task, or a trajectory to the corpus. Across 7 public agent benchmarks, the trained models gain 3.4 to 31.9 points of end-to-end success over prompting and, on all of them, reach the level of proprietary models. The gains concentrate where the supervision aimed, in repairs after a failed state. 
    \section{Related Work}
    \textbf{Decomposition and its unit.} Work on task decomposition sits between two poles that differ in what a part of the plan is. At one pole, least-to-most prompting and its successors elicit subproblems in free text, so a part is a sentence whose correctness nobody checks \citep{zhou2022least,wang2023plan}. Plans built this way cannot be trusted without outside verification \citep{kambhampati2024llms}. At the other pole, language models write PDDL for a symbolic planner, so a part is a predicate a solver decides \citep{liu2023llm+,guan2023leveraging}. The price is a symbol set a person must formalize for every new domain. \textit{SDP} keeps the part decidable and drops the formalization, since it writes each part as a state in natural language and decides it against observation at run time~\citep{jeong2026state}. The model that writes these states is prompted in \textit{SDP}. This paper trains it on the repair call.

    \textbf{Supervision for agents.} Agent supervision has scaled by collecting more of what worked. Tuning corpora collect interaction traces from capable models, so what they teach is how a stronger model behaves \citep{zeng2024agenttuning,chen2023fireact}. Synthesis pipelines replay tutorials or explore applications to manufacture such trajectories at scale \citep{xu2025agenttrek,sun2025genesis}. The environments themselves are now generated by program, from PDDL domains \citep{hu2025agentgen} to tool backends \citep{song2026envscaler} and full web sites \citep{guo2025genenv}. A generated environment is checked for being well formed and executable before agents run in it. Verification of the trajectories has tightened alongside, with execution checks and rule-based validation filtering what enters training \citep{liu2024apigen}. What survives into training is still the path that succeeded. Rejection sampling states this bias openly, since it keeps the completions whose answer passed and discards the rest \citep{zelikman2022star,yuan2023scaling}. This line therefore does not supply the unit repair needs, which is a verified failure paired with the repair that resolved it. Our worlds are also admitted before use, each with a goal reachable from its initial state. What we keep from them is the opposite record, the refusals their rules issued and the repairs that carried the run to a state the world certified.
        
    \textbf{Learning to repair.} Repair entered the literature as a prompting strategy. Reflexion turns a failure into a verbal lesson replayed in context, Self-Refine has the model critique its own output, and CRITIC routes interpreter and test results back into the prompt \citep{shinn2023reflexion,madaan2023self,gou2024critic}. Prompting has a documented limit, since without external feedback a model rarely locates its own fault and often revises an answer that was right \citep{huang2024large}. The response has been to train the ability, and training so far has meant reward. SCoRe reports that supervised fine-tuning on the model's own correction traces collapses onto the first attempt, and it therefore trains through multi-turn reinforcement learning \citep{kumar2025training}. Verifiable-reward and process-reward methods keep a verifier but reduce its verdict to a score, so the model learns that a step failed and never sees what the verifier found in its place \citep{lambert2024tulu,lightman2024let}. The collapse SCoRe reports was observed on corrections the model wrote for itself, judged only by the final answer. Our pairs differ on both counts. Each repair is conditioned on the refutation, which is evidence the first attempt never had, and a repair that restates the refuted chain is dropped at corpus construction. Regression onto the first attempt is therefore a pattern the training data never contains, and on such pairs repair is learnable by supervised fine-tuning alone.
    \section{Preliminaries: the state-centric runtime}
    \label{sec:prelim}

    This paper builds on the State-Centric Decision Process (\textit{SDP})~\citep{jeong2026state}, a runtime that makes every step of an agent's execution verifiable. An \textit{SDP} agent does not select actions against raw observations but commits to states of the world. Each state is written as a predicate, which is a natural-language condition that can be checked true or false. The task's goal is itself a predicate, written $g$. Four operators drive the runtime. \textsc{Propose} writes the plan, \textsc{Realize} acts toward it, \textsc{Validate} judges each state against the world, and \textsc{Replan} rewrites the plan after a state fails. \textsc{Propose} and \textsc{Replan} decide what the agent attempts, and this paper concerns those two. The other two stay as \textit{SDP} defines them, and one fact about \textsc{Validate} matters throughout. When a state fails, \textsc{Validate} records the observation that stood in its place. That observation describes what held and carries no diagnosis or advice.

    \textsc{Propose} is where an open task takes form. Given a goal that says only how the world should end up, it writes the states the world must pass through on the way. In \textit{SDP}, this is a prompted call to a large language model,
    \begin{equation}
        \hat{C} = \textsc{Propose}(g,\ c,\ \tau), \qquad \hat{C} = (\hat{s}_1, \ldots, \hat{s}_n), \qquad \hat{s}_n = g,
        \label{eq:interface}
    \end{equation}
    where $g$ is the goal, $c$ is the prefix of states the world has already certified, and $\tau$ is the refutation defined below, empty on the first call. The hat marks a state the world has not yet ruled on, and the runtime rules on one state at a time. \textsc{Realize} acts until it can report an observation for the first proposed state, and \textsc{Validate} decides that state against the observation,
    \begin{equation}
        o \leftarrow \textsc{Realize}(c,\ \hat{s}_1), \qquad
        (c,\ \tau) \leftarrow
        \begin{cases}
            (c \oplus \hat{s}_1,\ \varnothing) & \text{if } \textsc{Validate}(\hat{s}_1,\ o) \text{ certifies } \hat{s}_1,\\
            (c,\ (\hat{s}_1,\ o)) & \text{otherwise}.
        \end{cases}
        \label{eq:loop}
    \end{equation}
    A certified state joins the prefix and the chain advances to the next. When \textsc{Validate} refutes a state, that state and the observation recorded in its place become the refutation, written $\tau = (s_f,\ o_f)$. After a refutation Eq.~\ref{eq:interface} is called again from the current prefix, so execution moves forward only, the prefix is never reopened, and the loop ends when $g$ is certified. On the first call $\tau$ is empty and the call decomposes, and with $\tau$ present it repairs, the call \textit{SDP} names \textsc{Replan}. The two are one operator under the single interface of Eq.~\ref{eq:interface}, distinguished only by whether $\tau$ is present, and we call that operator the \emph{proposer}. Throughout, the lowercase nouns proposer, realizer, and validator name the models that fill \textsc{Propose}, \textsc{Realize}, and \textsc{Validate}.
    
    The model that fills this interface was never trained for it. What the runtime needs from the call is the chain whose execution most likely carries the world from where it stands to the goal,
    \begin{equation}
        \arg\max_{\hat{C}} \; P\big(g \text{ certified} \mid c,\ \hat{C}\big),
        \label{eq:obj}
    \end{equation}
    where the probability is over the runtime's stochastic execution of the chain through Eq.~\ref{eq:loop}, yet no prompted model was trained toward Eq.~\ref{eq:obj}. Its chains are whatever the prompt elicits, and they change when the wording does. The gap is widest on the calls that follow a refutation, for repairing under $\tau$ settles two decisions at once, what the rewritten remainder must pass through and how finely to cut it.
    \section{World State Generator}
    We build synthetic worlds whose rules a program enforces and whose goals are reachable from their initial states. Inside these worlds we run the \textit{SDP} loop and collect pairs of an input $(g,\ c,\ \tau)$ and the repair written after it, keeping the repairs that carried the run to a state the world certified. On these pairs we train the proposer for the repair call and call the trained model the World State Generator.  Its supervision is a pair, an input $(g,\ c,\ \tau)$ with a chain written after $\tau$ that carried the run to a state the world certified, and such pairs come only from worlds whose rules are enforced and whose goals are reachable from their initial states. This is why we build those worlds, run the \textit{SDP} loop inside them, and keep the repairs that carried the run to a state the world certified.

\begin{figure}[h]
    \centering
    \includegraphics[width=\textwidth]{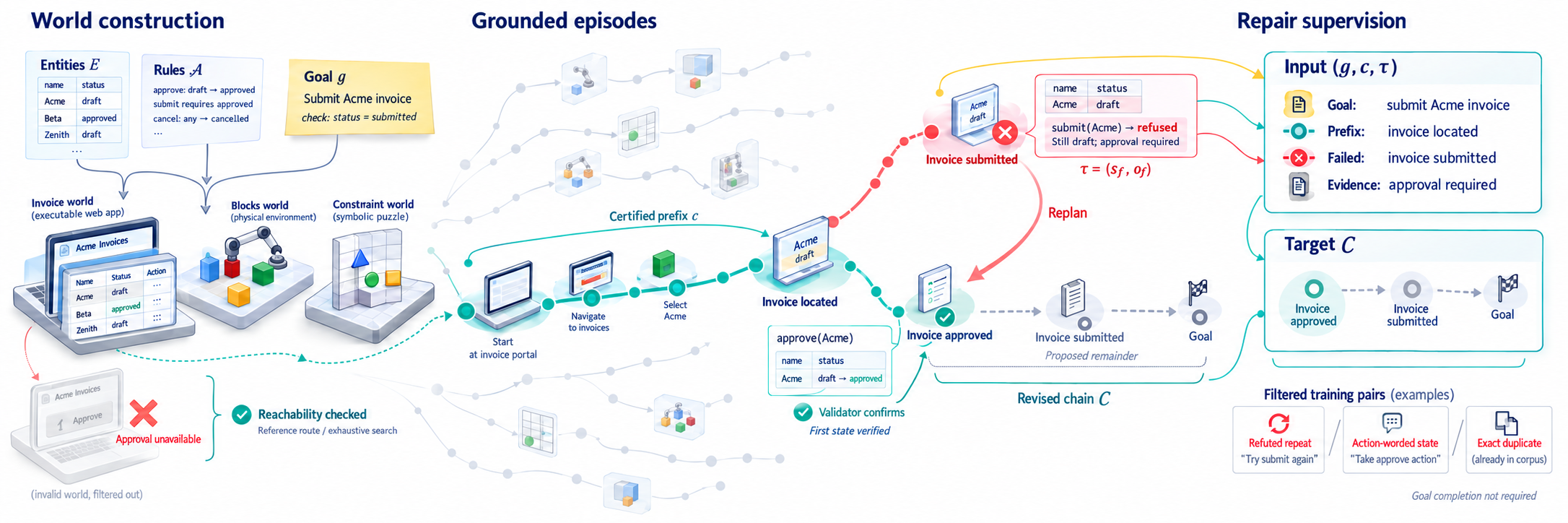}
    \caption{Left: worlds are built from entities, rules, and a goal, and a world is kept only if its goal is reachable from its initial state. Middle: agents run the \textit{SDP} loop inside these worlds. When the world refutes a state, the certified prefix stays and the rest of the chain is rewritten. Right: each refutation becomes a training pair, the input $(g, c, \tau)$ and the rewritten chain. A pair is kept when the rewritten chain reaches a state the world certifies, whether or not the run later reaches the goal.}
    \label{fig:extraction}
\end{figure}

    \subsection{The World as a State Chain}
        \label{sec:corpus}

        A world is a self-contained environment with three parts, \textbf{entities} with attributes, \textbf{actions} with rules, and a \textbf{goal} given as a sentence,
        \begin{equation}
            W = (E,\ A,\ g), \qquad g = (\ell_g,\ \chi_g).
            \label{eq:world}
        \end{equation}
        The attributes of $E$ are what an action in $A$ can read or change, and the rules of $A$ decide what each action does and when it is refused. The goal carries two forms, the sentence $\ell_g$ that the agent reads and the check $\chi_g$ that the program runs on the entities to decide whether the goal holds. In a tool world the entities are database rows such as invoices and tickets, the actions are tools that read and change them, a rule refuses a call on an invoice that is still in draft or on an id the table does not hold, and $\ell_g$ reads ``Put the invoice Acme Consulting through for payment'' while $\chi_g$ inspects the status of that row in the final table. Table~\ref{tab:worlds} gives the three parts for each of the 7 domains.

\begin{table*}[t]
    \centering
    \small
    \setlength{\tabcolsep}{5pt}
    \renewcommand{\arraystretch}{1.05}

    \begin{tabularx}{\textwidth}{
        @{} l l >{\raggedright\arraybackslash}X @{}
    }
        \toprule
        \textbf{Domain} & \textbf{Component} & \textbf{Description} \\
        \midrule

        \textbf{Physical}
            & Entities
            & Objects with locations and properties. \\
            & Actions
            & Primitive actions such as open, move, and combine. \\
            & Enforced rule
            & An action whose precondition does not hold changes nothing. \\
            & Goal check
            & The goal predicate on the final configuration. \\
            & Admission
            & Exhaustive search over configurations. \\
        \midrule

        \textbf{Software}
            & Entities
            & An index of Wikipedia documents. \\
            & Actions
            & Search, read, and note. \\
            & Enforced rule
            & A read of a title the index does not hold returns an error with similar titles, and a search with no terms is refused. \\
            & Goal check
            & The submitted answer against the gold answer. \\
            & Admission
            & Every gold document is found by a search on its title. \\
        \midrule

        \textbf{Web search}
            & Entities
            & Pages of a crawled snapshot. \\
            & Actions
            & Search, navigate, read, and note. \\
            & Enforced rule
            & A page opens only from a search result or a link on an open page. \\
            & Goal check
            & The submitted answer against the snapshot. \\
            & Admission
            & The reference route reaches the page, and exactly one page fits. \\
        \midrule

        \textbf{Tool use}
            & Entities
            & Database rows such as invoices and tickets. \\
            & Actions
            & 22 to 36 tools that search, list, read, create, update, and advance rows. \\
            & Enforced rule
            & A call on the wrong status or an unknown id is refused,
              and lists return ten rows per page. \\
            & Goal check
            & Row status and events in the final table. \\
            & Admission
            & The reference route is accepted, and a wrong run is rejected. \\
        \midrule

        \textbf{Constraint}
            & Entities
            & Candidates per slot with prices. \\
            & Actions
            & Inspect and select candidates. \\
            & Enforced rule
            & A selection that breaks a hidden policy is refused. \\
            & Goal check
            & Every constraint on the submitted selection. \\
            & Admission
            & Exhaustive search over combinations finds exactly one solution,
              and a decoy is rejected. \\
        \midrule

        \textbf{Reasoning}
            & Entities
            & A subject text and hidden facts. \\
            & Actions
            & Record a finding on an aspect, and conclude a verdict. \\
            & Enforced rule
            & A finding is checked against the hidden facts by program. \\
            & Goal check
            & The verdict against the hidden facts. \\
            & Admission
            & The verdict from the subject alone differs
              from the verdict with the facts. \\
        \midrule

        \textbf{Math}
            & Entities
            & A problem and the values determined so far. \\
            & Actions
            & Derive a value, and submit an answer. \\
            & Enforced rule
            & Only the submitted answer is checked,
              by program against the reference answer. \\
            & Goal check
            & The answer under tolerance. \\
            & Admission
            & The reference answer is machine checkable. \\

        \bottomrule
    \end{tabularx}

    \caption{%
        Entities, actions, enforced rules, goal checks,
        and admission criteria for the 7 domains.
    }
    \label{tab:worlds}
\end{table*}

        A world is composed by drawing its three parts and joining them, and the parts take the two sides a plan must cross, $\ell_g$ as the sentence the plan is written from and $E$ and $A$ as the world that holds the limits, dependencies, and orders the sentence never states. Joining the parts is not enough on its own, because the three are composed separately and a goal can demand a condition that $E$ and $A$ together never allow. A plan in such a world fails for no reason it could repair, so admission removes the world before any agent enters. Writing $e_0$ for the initial configuration of the entities, admission proves
        \begin{equation}
            \exists\, a_1, \ldots, a_k \in A \quad \text{such that} \quad \chi_g\big(a_k \circ \cdots \circ a_1(e_0)\big) = 1,
            \label{eq:admission}
        \end{equation}
        by executing a reference route where one can be written and by exhaustive search where it cannot. What remains is a world whose goal was reachable when the run began.

        Inside an admitted world the agent runs the \textit{SDP} loop, and one refutation from the invoice world shows what the loop records. The proposer writes a chain such as ``the invoice Acme Consulting has been located'', ``the invoice has been submitted for payment'', and the second state fails when the submit tool refuses an invoice that is still in draft. The validator reads the program's record of the call and the table and writes that the invoice is still in draft and that the submission was refused. That sentence is $o_f$, and with the refuted state it is the $\tau$ of Eq.~\ref{eq:loop}. From the table as the refused call left it, the proposer places ``the invoice has been approved'' before the submission and keeps the rest.

        The invoice run divides its decisions in a way that holds in every domain. What a call returns, what an action changes, and whether $\chi_g$ holds are decided by the program. Whether an intermediate state holds is decided by the validator, a language model that reads the program's record and answers only that question, except in reasoning worlds, where a program checks each finding against the hidden facts. An episode ends when $\chi_g$ passes or when the attempt budget is spent, and a chain that completes without $\chi_g$ passing is handled by domain. In physical, software, and web worlds the check returns only that the goal does not hold, so there is nothing to repair from and the episode fails, while in tool, constraint, reasoning, and math worlds the check can name the condition it found unmet or the answer it rejected, and the proposer receives that as one more refutation and repairs from it.

        A run leaves the world written as a chain of states, and the chains written after a refutation are what the corpus keeps,
        \begin{equation}
            \mathcal{D} = \big\{\, (x,\ C) \;:\; x = (g,\ c,\ \tau),\ \ C = (s_1, \ldots, s_n),\ \ s_n = g,\ \ \exists\, i \le n \text{ such that the run certified } s_i \,\big\},
            \label{eq:corpus}
        \end{equation}
        one repair for every $\tau$ after which the run certified at least one state of the rewritten chain. From the invoice episode Eq.~\ref{eq:corpus} yields one example, with the goal, the prefix up to ``located'', and the $\tau$ of the refused submission as $x$, and the chain from ``approved'' to the goal as $C$. The target is the whole remainder the proposer wrote, and what is certified about it is the prefix the run went on to reach, at least the decision of what to establish first where the previous chain broke, while the rest of the chain is decomposition that the run may or may not have carried out. 48.9\% of the kept repairs come from episodes that went on to certify the goal, and the rest from episodes that certified part of the repair but ended short of the goal. A repair is dropped if it restates, as a set of states, a chain that was already refuted from the same prefix, if any of its states names an action rather than a condition, or if it duplicates a kept example exactly.

        Table~\ref{tab:corpus} gives the yield of Eq.~\ref{eq:corpus}. Three open models near 30B parameters, Qwen3-32B, Qwen3.6-27B, and Gemma4-31B, ran the loop over the same worlds for 226,316 episodes, and the world refuted at least one state in 75.3\% of them. Those refutations produced 775,180 repair calls, and 238,563 of the repairs, 30.8\%, passed the filters. The domains contribute unequally, from 75,455 repairs in physical worlds to 12,243 in tool worlds. The difference comes first from how often each world refutes, 253,286 repair calls in physical worlds against 69,094 in tool worlds, and only second from the share that passes, which ranges from 17.7\% in tool worlds to 50.0\% in reasoning worlds. A kept repair holds 5.5 states on average, from 1.2 in constraint worlds to 11.2 in physical worlds, which is shorter than the 5.9 states of a first plan because the certified prefix is not rewritten. An episode attempts 8.7 states on average, from 4.8 in web worlds to 18.5 in physical worlds.

\begin{table}[t]
    \centering
    \small
    \resizebox{\linewidth}{!}{%
    \begin{tabular}{l l l r r r r r}
        \toprule
        \textbf{Domain} & \textbf{What the program decides} & \textbf{Model} & \textbf{Episodes} & \textbf{Refuted (\%)} & \textbf{Reached (\%)} & \textbf{Repair calls} & \textbf{Kept (\%)} \\
        \midrule
        \multirow{3}{*}{Physical} & \multirow{3}{*}{the goal check on the engine state} & Qwen3-32B & 11,064 & 90.6 & 24.9 & 85,921 & 33.3 \\
         &  & Qwen3.6-27B & 10,614 & 91.5 & 19.1 & 87,231 & 39.2 \\
         &  & Gemma4-31B & 10,097 & 87.9 & 19.0 & 80,134 & 15.8 \\
        \midrule
        \multirow{3}{*}{Software} & \multirow{3}{*}{the submitted answer against the gold answer} & Qwen3-32B & 11,040 & 86.2 & 43.4 & 37,870 & 27.0 \\
         &  & Qwen3.6-27B & 10,239 & 69.4 & 60.6 & 27,357 & 30.7 \\
         &  & Gemma4-31B & 10,380 & 67.3 & 53.0 & 28,200 & 33.3 \\
        \midrule
        \multirow{3}{*}{Web search} & \multirow{3}{*}{the answer against the snapshot} & Qwen3-32B & 13,338 & 51.3 & 87.6 & 19,671 & 37.7 \\
         &  & Qwen3.6-27B & 10,306 & 39.9 & 90.5 & 14,858 & 29.3 \\
         &  & Gemma4-31B & 11,279 & 52.6 & 72.7 & 25,375 & 21.8 \\
        \midrule
        \multirow{3}{*}{Tool use} & \multirow{3}{*}{the database after the run} & Qwen3-32B & 12,614 & 73.9 & 78.5 & 29,272 & 16.0 \\
         &  & Qwen3.6-27B & 10,141 & 78.3 & 88.8 & 22,359 & 20.1 \\
         &  & Gemma4-31B & 11,074 & 67.7 & 94.6 & 17,463 & 17.5 \\
        \midrule
        \multirow{3}{*}{Constraint} & \multirow{3}{*}{every constraint on the submitted selection} & Qwen3-32B & 11,345 & 97.9 & 39.3 & 46,873 & 17.3 \\
         &  & Qwen3.6-27B & 10,495 & 96.0 & 54.9 & 35,979 & 20.1 \\
         &  & Gemma4-31B & 10,817 & 97.6 & 62.5 & 26,208 & 34.6 \\
        \midrule
        \multirow{3}{*}{Reasoning} & \multirow{3}{*}{the verdict against the hidden facts} & Qwen3-32B & 10,508 & 96.8 & 53.2 & 55,481 & 43.2 \\
         &  & Qwen3.6-27B & 10,444 & 87.3 & 78.6 & 34,922 & 58.9 \\
         &  & Gemma4-31B & 10,003 & 75.2 & 79.6 & 27,922 & 52.5 \\
        \midrule
        \multirow{3}{*}{Math} & \multirow{3}{*}{the final answer under tolerance} & Qwen3-32B & 10,291 & 58.3 & 61.2 & 22,950 & 31.0 \\
         &  & Qwen3.6-27B & 10,203 & 71.6 & 55.1 & 30,621 & 31.9 \\
         &  & Gemma4-31B & 10,024 & 46.4 & 66.4 & 18,513 & 27.4 \\
        \midrule
        \multicolumn{3}{l}{Total} & 226,316 & 75.3 & 61.5 & 775,180 & 30.8 \\
        \bottomrule
    \end{tabular}%
    }
    \caption{7 domains, what the program grades in each, and what each extraction model produced inside them. Episodes counts runs of the \textit{SDP} loop, Refuted the share of episodes in which the world refuted at least one state, and Reached the share that certified the goal. Repair calls counts the repair calls the proposer made, and Kept the share of those that survived the corpus filters. Intermediate states are judged by the validator, by a program in reasoning worlds, and the goal by the program in every domain.}
    \label{tab:corpus}
\end{table}

    \subsection{Training the World State Generator}
        \label{sec:training}

        A training example is one element of $\mathcal{D}$ in Eq.~\ref{eq:corpus}, serialized with the prompt the proposer received at extraction, so the training input is the call itself. \textit{WSG} is the backbone with LoRA~\citep{hu2021lora} adapters $\theta$ on every attention and feed-forward projection of its language model, and $\theta$ is trained by supervised fine-tuning on these pairs alone,
        \begin{equation}
            \mathcal{L}(\theta) = - \sum_{(x,\ C) \in \mathcal{D}} \ \sum_{t=1}^{|C|} \log p_\theta\big(C_t \mid x,\ C_{<t}\big),
            \label{eq:sft}
        \end{equation}
        where $C_t$ ranges over the tokens of the chain and the tokens of $x$ carry no loss. No reward and no comparison between chains enters the objective. Every chain in $\mathcal{D}$ holds a state the world certified after the previous chain broke, so Eq.~\ref{eq:sft} raises the likelihood of remainders the world certified at least in part, and a remainder the world refused at every state never enters the objective.

        The 7 domains contribute unequally, so examples are drawn per domain with probability
        \begin{equation}
            p_d \ \propto\ n_d^{\,1/2},
            \label{eq:mix}
        \end{equation}
        where $n_d$ is the number of examples the domain holds. Eq.~\ref{eq:mix} keeps the largest domains from crowding out the smallest without discarding their examples, and both backbones we train use the same corpus, objective, and mixture.

        Two properties of this supervision let it carry beyond the worlds it came from. Each target is conditioned on $\tau$, the evidence the first attempt did not have, so the model learns to read a refutation and write the remainder from it rather than to write chains from the goal alone. The states it learns to write are free descriptions of conditions, not symbols from a fixed vocabulary, so what the model learns is a map from refutation to remainder over descriptions, and nothing in that map names the tools, pages, or objects of a particular world.
    \section{Experiments}
    We train \textit{WSG} on the corpus of Section~\ref{sec:corpus} with Eq.~\ref{eq:sft}, on two open backbones, Gemma4-31B~\citep{team2026gemma} and Qwen3.6-27B~\citep{qwen36_27b}, since smaller backbones had difficulty writing a full chain from a long context. We evaluate zero-shot on 7 public agent benchmarks, $\tau^2$-bench~\citep{barres2025tau}, ScienceWorld~\citep{wang2022scienceworld}, ToolSandbox~\citep{lu2025toolsandbox}, PlanBench-XL~\citep{liu2026planbench}, REAL~\citep{garg2026real}, DeepPlanning~\citep{zhang2026deepplanning}, and AgentBoard's PDDL~\citep{ma2024agentboard}. Every run uses the \textit{SDP} runtime, and in every run the same backbone fills the realizer and the validator. In the \textit{SDP} condition the backbone is prompted for every proposer call, and in the \textit{WSG} condition the same backbone writes the first plan while \textit{WSG} takes every repair call, with budgets and decoding held identical. Each benchmark is scored by its own metric, which Table~\ref{tab:main} lists. An aborted episode counts as a failure, and every number is the mean over the benchmark's task set.

    \subsection{Training}        
        Both backbones train on the same corpus with the same settings. The corpus holds 238,563 repair examples, and 2\% of the worlds are held out with every example they produced. The settings are LoRA rank 32 and alpha 64, learning rate $10^{-4}$ with cosine decay, sequences packed to 4,096 tokens, an effective batch of 96, and 10 epochs in bf16. Each run uses two H200 GPUs and takes 12 to 16 hours per epoch. The loss on held-out worlds falls monotonically on both backbones and flattens within the first 1,200 steps, and no gap opens between training and held-out loss over the remaining epochs. Every chain the model generates from held-out repair prompts parses as a valid state list from the first evaluation onward.

        When the training set included first plans, written without $\tau$, the model overfit faster than on repairs alone and lost the planning it had under prompting. A first plan is conditioned on the goal alone, so it teaches the shape of a chain rather than a response to evidence, and a model trained on shapes overwrites its habit of planning instead of adding a conditional one. A repair does not do this, because it is tied to the $\tau$ that produced it. The same collapse appeared with a larger adapter or a higher learning rate, and it was sharpest when one domain dominated the mixture. The adapter is small, the learning rate is low, and Eq.~\ref{eq:mix} balances the domains for this reason.

        The corpus keeps the repairs each backbone extracted for itself alongside those of the other two models, because runs trained without a backbone's own repairs recovered less often on held-out worlds. The corpus is large and spans 7 domains because a repair is specific to the rule that was broken and the rules differ by world, so a model learns to read a refutation rather than memorize one world's fixes only after seeing refutations of many kinds.
        
\begin{table}[h]
    \centering
    \small
    \setlength{\tabcolsep}{4pt}
    \newcommand{\prop}[2]{\begin{tabular}[c]{@{}c@{}}#1\\[-2pt]{\scriptsize #2}\end{tabular}}
    \resizebox{\linewidth}{!}{%
    \begin{tabular}{l l cccc cc cc}
        \toprule
        & & & & & & \multicolumn{2}{c}{\textbf{Gemma4-31B}} & \multicolumn{2}{c}{\textbf{Qwen3.6-27B}} \\
        \cmidrule(lr){7-8} \cmidrule(lr){9-10}
        \textbf{Benchmark} & \textbf{Metric} & \multicolumn{4}{c}{\textbf{Proprietary models (published)}} & SDP & WSG & SDP & WSG \\
        \midrule
        $\tau^2$-bench      & Accuracy      & \prop{46.3}{Gemini-2.5-Pro}  & \prop{53.7}{Sonnet-4}         & \prop{54.3}{GPT-4.1}             & \prop{\textbf{79.9}}{GPT-5-Think}     & 42.5 & 56.3 {\scriptsize(+13.8)} & 68.3 & \textbf{76.6} {\scriptsize(+8.3)} \\
        ScienceWorld        & Average score & \prop{36.4}{GPT-4 ReAct}     & \prop{39.2}{GPT-4 CoT}        & \prop{45.3}{GPT-4 Reflexion}     & \prop{\textbf{47.9}}{GPT-4 Plan-and-Act} & 35.8 & 44.4 {\scriptsize(+8.6)} & 36.4 & \textbf{46.8} {\scriptsize(+10.4)} \\
        ToolSandbox         & Average score & \prop{64.3}{GPT-4}           & \prop{65.6}{GPT-3.5}          & \prop{69.2}{Opus-3}              & \prop{\textbf{73.0}}{GPT-4o}          & 64.3 & \textbf{72.5} {\scriptsize(+8.2)} & 64.7 & 69.6 {\scriptsize(+4.9)} \\
        PlanBench-XL        & Accuracy      & \prop{51.9}{GPT-5.4}         & \prop{52.2}{Gemini-3.5-Flash} & \prop{63.1}{DeepSeek-V4-Flash}   & \prop{\textbf{77.1}}{Gemini-3.1-Pro}  & 38.5 & \textbf{53.9} {\scriptsize(+15.4)} & 16.1 & 48.0 {\scriptsize(+31.9)} \\
        REAL                & Accuracy      & \prop{25.0}{o3-mini}         & \prop{35.0}{Gemini-2.5-Pro}   & \prop{38.0}{o3-pro}              & \prop{\textbf{41.0}}{Sonnet-3.7}      & 27.7 & 37.3 {\scriptsize(+9.6)} & 22.6 & \textbf{40.2} {\scriptsize(+17.6)} \\
        DeepPlanning (Shop) & Match score   & \prop{50.1}{Grok-4.1-Fast}   & \prop{58.6}{GPT-5.2}          & \prop{69.1}{Gemini-2.5-Pro}      & \prop{\textbf{75.8}}{Sonnet-4.5}      & 48.5 & \textbf{59.0} {\scriptsize(+10.5)} & 29.1 & 51.2 {\scriptsize(+22.1)} \\
        AgentBoard (PDDL)   & Success rate  & \prop{13.3}{Haiku-3}         & \prop{40.0}{Claude-2}         & \prop{61.7}{GPT-4o}              & \prop{\textbf{61.7}}{GPT-4}           & 53.3 & 58.3 {\scriptsize(+5.0)} & 58.3 & \textbf{61.7} {\scriptsize(+3.4)} \\
        \bottomrule
    \end{tabular}%
    }
    \caption{Zero-shot performance on 7 agent benchmarks. SDP prompts the backbone for every proposer call, and WSG replaces the calls under $\tau$ with the trained model on the same backbone and runtime, with the gain over SDP in parentheses. \textbf{Bold} marks the best proprietary result and the best open result per row.}
    \label{tab:main}
\end{table}

    \subsection{Zero-shot Performance on Agent Benchmarks}
        \textit{WSG} raises performance across all 7 benchmarks (Table~\ref{tab:main}). The gain is largest where the prompted proposer is weakest, 32 points for Qwen3.6-27B on PlanBench-XL, and smallest on AgentBoard, where both backbones already succeed on more than half the tasks before training. On every benchmark \textit{WSG} lands within the range of published proprietary results, and on five of them within a few points of the best, while on PlanBench-XL and DeepPlanning the strongest published systems remain well ahead. Much of this gap comes from the realizer and the validator, which we discuss in Section~\ref{sec:discussion}.
        
        Every point of this gain comes from what the agent does after the world has refuted a state, since the two conditions differ only in the calls made under $\tau$, and the logs show what changed there. When the prompted proposer receives a refutation, it often rewrites the refuted state in different words, in 38 percent of repairs on REAL and 28 percent on ScienceWorld for Gemma4-31B, and the rest of its chain changes with it. \textit{WSG} keeps the refuted state and changes what surrounds it, rewording it in 19 and 7 percent of repairs on the same benchmarks, and its rewritten chain overlaps the chain it replaced far more, 0.49 against 0.29 in state overlap. What it inserts is the condition the refutation named. On ToolSandbox, for example, the observation reports a tool call rejected for a parameter below its valid range, the prompted proposer concludes that no reminders exist and closes the task, and \textit{WSG} drops the filter, retrieves the full list, and reaches the goal. The difference is where each proposer looks for the fix. The prompted proposer searches the language of the plan, and \textit{WSG} reads the rule the world has just stated, a parameter range, a button that must be enabled, a room the agent must first be in, and bends the plan to it.
        
    \subsection{Recovery after Refutation and Its Cost}        
        \begin{table}[h]
            \centering
            \small
            \setlength{\tabcolsep}{5pt}
            \resizebox{\linewidth}{!}{%
            \begin{tabular}{l cccc cccc}
                \toprule
                & \multicolumn{4}{c}{\textbf{Gemma4-31B}} & \multicolumn{4}{c}{\textbf{Qwen3.6-27B}} \\
                \cmidrule(lr){2-5} \cmidrule(lr){6-9}
                \textbf{Benchmark} & Recovery (\%) & Steps & Replans & Time (s) & Recovery (\%) & Steps & Replans & Time (s) \\
                \midrule
                $\tau^2$-bench      & 51.0 / 64.0 & 42.2 / 41.8 & 5.91 / 5.57 & 15.6 / 13.9 & 74.0 / 81.0 & 38.8 / 42.6 & 3.57 / 2.47 & 16.3 / 15.6 \\
                ScienceWorld        & 10.6 / 27.5 & 73.6 / 43.8 & 4.19 / 3.07 & 348 / 223   & 13.8 / 22.9 & 73.1 / 40.2 & 3.89 / 2.23 & 404 / 215 \\
                ToolSandbox         & 71.3 / 85.1 & 10.9 / 10.1 & 2.14 / 1.05 & 27 / 31     & 66.5 / 72.7 & 11.9 / 12.1 & 2.37 / 1.46 & 22 / 20 \\
                PlanBench-XL        & 40.2 / 63.1 & 32.0 / 30.7 & 1.90 / 4.55 & 307 / 335   & 20.8 / 48.4 & 32.8 / 36.2 & 1.30 / 4.26 & 259 / 353 \\
                REAL                & 17.2 / 25.0 & 33.6 / 27.6 & 3.94 / 2.43 & 832 / 376   & 18.9 / 23.0 & 41.9 / 28.6 & 3.92 / 2.01 & 1077 / 478 \\
                DeepPlanning (Shop) & 8.7 / 17.2  & 55.6 / 67.3 & 6.74 / 4.26 & 2116 / 1036 & 6.8 / 19.5  & 30.8 / 50.0 & 2.92 / 2.73 & 1223 / 1511 \\
                AgentBoard (PDDL)   & 20.0 / 26.7 & 57.2 / 58.1 & 4.26 / 3.65 & 455 / 345   & 24.1 / 25.9 & 53.2 / 51.5 & 3.13 / 2.10 & 551 / 335 \\
                \bottomrule
            \end{tabular}%
            }
            \caption{Recovery after refutation and its cost, reported as SDP / WSG. Recovery is the share of episodes that reached the goal after at least one refutation. Steps counts messages per episode, Replans counts proposer calls under $\tau$ per episode, and Time is wall-clock seconds per episode.}
            \label{tab:recovery}
        \end{table}
        
        \textit{WSG} adds the most recovery where the prompted proposer recovered least, and Table~\ref{tab:recovery} shows this on the refuted episodes behind the gain in Table~\ref{tab:main}. Recovery is the share of refuted episodes that still reached the goal. It can exceed the success rate in Table~\ref{tab:main} because an episode the world never refutes can still end on a wrong answer that only the benchmark's final check rejects. On DeepPlanning and ScienceWorld the prompted proposer recovered fewer than one refuted episode in seven, and \textit{WSG} raises that share by 66\% to 187\%. On $\tau^2$-bench and ToolSandbox the prompted proposer already recovered more than half of its refuted episodes, and the rise there stays between 9\% and 25\%. Training on refutations adds the most where prompting could not repair a refusal.

        What that recovery costs depends on where the prompted proposer was failing. On 6 of the 7 benchmarks replans per episode fall as recovery rises, because the calls the prompted proposer spent rewording a refuted state are no longer made. Wall-clock time falls where those calls were most expensive, by 55\% on REAL with Gemma4-31B and by 47\% on ScienceWorld with Qwen3.6-27B. PlanBench-XL is the exception, and there replans more than double on Gemma4-31B and more than triple on Qwen3.6-27B, because the prompted proposer stops after fewer calls where \textit{WSG} keeps rewriting, and those added calls are what buy the largest point rise in recovery in the table. Recovery is cheaper where the prompted proposer wasted calls and dearer where it made too few.
    \section{Discussion}
    \label{sec:discussion}

    Most remaining failures come from the realizer and the validator, which \textit{WSG} leaves untrained. Training changes only the repair call, and the realizer and the validator stay prompted. Judged by hand on the failed episodes of DeepPlanning and ScienceWorld, the realizer fails to reach a correctly written state in 45\% of them. In another 20\% the validator refutes a state that already holds and triggers a repair the run did not need. Recovery on these two benchmarks stays below 30\% after training because these failures lie outside the repair call. The same two roles stay prompted on PlanBench-XL and DeepPlanning, where the gap to the strongest published systems is largest. Extending training to the realizer and the validator is therefore the next step.

    The worlds that produced the repair corpus also hold supervision for the other two roles. Every call the realizer made is logged with the program's record of whether it was refused and what it changed. These records can train the realizer to reach a written state. The validator needs a label for whether a state holds, and reasoning worlds already supply one by checking each finding against the hidden facts. Writing such a check for the states of the other domains would give the validator the same supervision. The worlds also support reinforcement learning, since the program's goal check defines a reward and the observation in $\tau$ gives feedback at the failed state. Whether learning from both signals raises recovery beyond supervised repair remains open.
    \section{Conclusion}
    We introduced World State Generator, a planner trained to keep its plan aligned with the world it runs in. When a failure reveals a constraint the plan overlooked, \textit{WSG} uses that evidence to revise the remaining plan while preserving the conditions already achieved. To train it, we built a pipeline that pairs failures with the repairs that followed them in 7 domains of synthetic worlds and collected about 226K trajectories. Each world is checked for goal reachability before agents run, and a repair is kept only when it leads to a state the world certified. Trained only on these worlds, two open backbones near 30B parameters improve end-to-end success over prompting on 7 public benchmarks and reach the level of proprietary models on all of them. The gains concentrate in recovery after failed states, where the supervision aimed. These results show that verified failure and repair pairs from diverse synthetic worlds can teach a planner to use failure evidence in unseen environments.

    \section*{Acknowledgement}
        This work was supported in part by the DARPA Young Faculty Award, the National Science Foundation (NSF) under Grants \#2127780, \#2319198, \#2321840, \#2312517, and \#2235472, the Semiconductor Research Corporation (SRC), the Office of Naval Research through the Young Investigator Program Award, and Grants \#N00014-21-1-2225 and N00014-24-1-2547, Army Research Office Grant \#W911NF2410360. Additionally, support was provided by the Air Force Office of Scientific Research under Award \#FA9550-22-1-0253.
    {
        \small
        \bibliographystyle{plain}
        \bibliography{ref.bib}

@article{jeong2026state,
  title={State-Centric Decision Process},
  author={Jeong, Sungheon and Masukawa, Ryozo and Yun, Sanggeon and Imani, Mahdi and Imani, Mohsen},
  journal={arXiv preprint arXiv:2605.12755},
  year={2026}
}

@inproceedings{hu2025agentgen,
  title={Agentgen: Enhancing planning abilities for large language model based agent via environment and task generation},
  author={Hu, Mengkang and Zhao, Pu and Xu, Can and Sun, Qingfeng and Lou, Jian-Guang and Lin, Qingwei and Luo, Ping and Rajmohan, Saravan},
  booktitle={Proceedings of the 31st ACM SIGKDD Conference on Knowledge Discovery and Data Mining V. 1},
  pages={496--507},
  year={2025}
}

@article{zhou2022least,
  title={Least-to-most prompting enables complex reasoning in large language models},
  author={Zhou, Denny and Sch{\"a}rli, Nathanael and Hou, Le and Wei, Jason and Scales, Nathan and Wang, Xuezhi and Schuurmans, Dale and Cui, Claire and Bousquet, Olivier and Le, Quoc and others},
  journal={arXiv preprint arXiv:2205.10625},
  year={2022}
}

@inproceedings{wang2023plan,
  title={Plan-and-solve prompting: Improving zero-shot chain-of-thought reasoning by large language models},
  author={Wang, Lei and Xu, Wanyu and Lan, Yihuai and Hu, Zhiqiang and Lan, Yunshi and Lee, Roy Ka-Wei and Lim, Ee-Peng},
  booktitle={Proceedings of the 61st annual meeting of the association for computational linguistics (volume 1: Long papers)},
  pages={2609--2634},
  year={2023}
}

@article{liu2023llm+,
  title={Llm+ p: Empowering large language models with optimal planning proficiency},
  author={Liu, Bo and Jiang, Yuqian and Zhang, Xiaohan and Liu, Qiang and Zhang, Shiqi and Biswas, Joydeep and Stone, Peter},
  journal={arXiv preprint arXiv:2304.11477},
  year={2023}
}

@article{guan2023leveraging,
  title={Leveraging pre-trained large language models to construct and utilize world models for model-based task planning},
  author={Guan, Lin and Valmeekam, Karthik and Sreedharan, Sarath and Kambhampati, Subbarao},
  journal={Advances in Neural Information Processing Systems},
  volume={36},
  pages={79081--79094},
  year={2023}
}

@article{kambhampati2024llms,
  title={Llms can't plan, but can help planning in llm-modulo frameworks},
  author={Kambhampati, Subbarao and Valmeekam, Karthik and Guan, Lin and Verma, Mudit and Stechly, Kaya and Bhambri, Siddhant and Saldyt, Lucas and Murthy, Anil},
  journal={arXiv preprint arXiv:2402.01817},
  year={2024}
}

@inproceedings{zeng2024agenttuning,
  title={Agenttuning: Enabling generalized agent abilities for llms},
  author={Zeng, Aohan and Liu, Mingdao and Lu, Rui and Wang, Bowen and Liu, Xiao and Dong, Yuxiao and Tang, Jie},
  booktitle={Findings of the Association for Computational Linguistics: ACL 2024},
  pages={3053--3077},
  year={2024}
}

@article{chen2023fireact,
  title={Fireact: Toward language agent fine-tuning},
  author={Chen, Baian and Shu, Chang and Shareghi, Ehsan and Collier, Nigel and Narasimhan, Karthik and Yao, Shunyu},
  journal={arXiv preprint arXiv:2310.05915},
  year={2023}
}

@inproceedings{xu2025agenttrek,
  title={Agenttrek: Agent trajectory synthesis via guiding replay with web tutorials},
  author={Xu, Yiheng and Lu, Dunjie and Shen, Zhennan and Wang, Junli and Wang, Zekun and Mao, Yuchen and Xiong, Caiming and Yu, Tao},
  booktitle={International Conference on Learning Representations},
  volume={2025},
  pages={79822--79843},
  year={2025}
}

@inproceedings{sun2025genesis,
  title={Os-genesis: Automating gui agent trajectory construction via reverse task synthesis},
  author={Sun, Qiushi and Cheng, Kanzhi and Ding, Zichen and Jin, Chuanyang and Wang, Yian and Xu, Fangzhi and Wu, Zhenyu and Jia, Chengyou and Chen, Liheng and Liu, Zhoumianze and others},
  booktitle={Proceedings of the 63rd Annual Meeting of the Association for Computational Linguistics (Volume 1: Long Papers)},
  pages={5555--5579},
  year={2025}
}

@inproceedings{song2026envscaler,
  title={Envscaler: Scaling tool-interactive environments for llm agent via programmatic synthesis},
  author={Song, Xiaoshuai and Chang, Haofei and Dong, Guanting and Zhu, Yutao and Wen, Ji-Rong and Dou, Zhicheng},
  booktitle={Findings of the Association for Computational Linguistics: ACL 2026},
  pages={8326--8357},
  year={2026}
}

@article{guo2025genenv,
  title={Genenv: Difficulty-aligned co-evolution between llm agents and environment simulators},
  author={Guo, Jiacheng and Yang, Ling and Chen, Peter and Xiao, Qixin and Wang, Yinjie and Juan, Xinzhe and Qiu, Jiahao and Shen, Ke and Wang, Mengdi},
  journal={arXiv preprint arXiv:2512.19682},
  year={2025}
}

@article{liu2024apigen,
  title={Apigen: Automated pipeline for generating verifiable and diverse function-calling datasets},
  author={Liu, Zuxin and Hoang, Thai and Zhang, Jianguo and Zhu, Ming and Lan, Tian and Kokane, Shirley and Tan, Juntao and Yao, Weiran and Liu, Zhiwei and Feng, Yihao and others},
  journal={Advances in Neural Information Processing Systems},
  volume={37},
  pages={54463--54482},
  year={2024}
}

@article{zelikman2022star,
  title={Star: Bootstrapping reasoning with reasoning},
  author={Zelikman, Eric and Wu, Yuhuai and Mu, Jesse and Goodman, Noah},
  journal={Advances in Neural Information Processing Systems},
  volume={35},
  pages={15476--15488},
  year={2022}
}

@article{yuan2023scaling,
  title={Scaling relationship on learning mathematical reasoning with large language models},
  author={Yuan, Zheng and Yuan, Hongyi and Li, Chengpeng and Dong, Guanting and Lu, Keming and Tan, Chuanqi and Zhou, Chang and Zhou, Jingren},
  journal={arXiv preprint arXiv:2308.01825},
  year={2023}
}

@article{shinn2023reflexion,
  title={Reflexion: Language agents with verbal reinforcement learning},
  author={Shinn, Noah and Cassano, Federico and Gopinath, Ashwin and Narasimhan, Karthik and Yao, Shunyu},
  journal={Advances in neural information processing systems},
  volume={36},
  pages={8634--8652},
  year={2023}
}

@article{madaan2023self,
  title={Self-refine: Iterative refinement with self-feedback},
  author={Madaan, Aman and Tandon, Niket and Gupta, Prakhar and Hallinan, Skyler and Gao, Luyu and Wiegreffe, Sarah and Alon, Uri and Dziri, Nouha and Prabhumoye, Shrimai and Yang, Yiming and others},
  journal={Advances in neural information processing systems},
  volume={36},
  pages={46534--46594},
  year={2023}
}

@inproceedings{gou2024critic,
  title={Critic: Large language models can self-correct with tool-interactive critiquing},
  author={Gou, Zhibin and Shao, Zhihong and Gong, Yeyun and Yang, Yujiu and Duan, Nan and Chen, Weizhu and others},
  booktitle={International Conference on Learning Representations},
  volume={2024},
  pages={57734--57811},
  year={2024}
}

@inproceedings{huang2024large,
  title={Large language models cannot self-correct reasoning yet},
  author={Huang, Jie and Chen, Xinyun and Mishra, Swaroop and Zheng, Huaixiu Steven and Yu, Adams and Song, Xinying and Zhou, Denny},
  booktitle={International conference on learning representations},
  volume={2024},
  pages={32808--32824},
  year={2024}
}

@inproceedings{kumar2025training,
  title={Training language models to self-correct via reinforcement learning},
  author={Kumar, Aviral and Zhuang, Vincent and Agarwal, Rishabh and Su, Yi and Co-Reyes, John D and Singh, Avi and Baumli, Kate and Iqbal, Shariq and Bishop, Colton and Roelofs, Rebecca and others},
  booktitle={International Conference on Learning Representations},
  volume={2025},
  pages={54523--54549},
  year={2025}
}

@article{lambert2024tulu,
  title={Tulu 3: Pushing frontiers in open language model post-training},
  author={Lambert, Nathan and Morrison, Jacob and Pyatkin, Valentina and Huang, Shengyi and Ivison, Hamish and Brahman, Faeze and Miranda, Lester James V and Liu, Alisa and Dziri, Nouha and Lyu, Shane and others},
  journal={arXiv preprint arXiv:2411.15124},
  year={2024}
}

@inproceedings{lightman2024let,
  title={Let's verify step by step},
  author={Lightman, Hunter and Kosaraju, Vineet and Burda, Yuri and Edwards, Harrison and Baker, Bowen and Lee, Teddy and Leike, Jan and Schulman, John and Sutskever, Ilya and Cobbe, Karl},
  booktitle={International Conference on Learning Representations},
  volume={2024},
  pages={39578--39601},
  year={2024}
}

@article{yan2026tide,
  title={Tide: Trajectory-based diagnostic evaluation of test-time improvement in llm agents},
  author={Yan, Hang and Che, Xinyu and Xu, Fangzhi and Sun, Qiushi and Ding, Zichen and Cheng, Kanzhi and Zhang, Jian and Qin, Tao and Liu, Jun and Lin, Qika},
  journal={arXiv preprint arXiv:2602.02196},
  year={2026}
}

@article{li2026benchmark,
  title={Benchmark test-time scaling of general llm agents},
  author={Li, Xiaochuan and Ming, Ryan and Setlur, Pranav and Paladugu, Abhijay and Tang, Andy and Kang, Hao and Shao, Shuai and Jin, Rong and Xiong, Chenyan},
  journal={arXiv preprint arXiv:2602.18998},
  year={2026}
}

@article{hu2021lora,
  title={Lora: Low-rank adaptation of large language models},
  author={Hu, Edward J and Shen, Yelong and Wallis, Phillip and Allen-Zhu, Zeyuan and Li, Yuanzhi and Wang, Shean and Wang, Lu and Chen, Weizhu},
  journal={arXiv preprint arXiv:2106.09685},
  year={2021}
}

@article{team2026gemma,
  title={Gemma 4 technical report},
  author={Team, Gemma and Abd, Sherif El and Aggarwal, Vaibhav and Algayres, Robin and Andreev, Alek and Bachem, Olivier and Ballantyne, Ian and Brick, Cormac and C{\u{a}}rbune, Victor and Casbon, Michelle and others},
  journal={arXiv preprint arXiv:2607.02770},
  year={2026}
}

@misc{qwen36_27b,
    title = {{Qwen3.6-27B}: Flagship-Level Coding in a 27B Dense Model},
    url = {https://qwen.ai/blog?id=qwen3.6-27b},
    author = {{Qwen Team}},
    month = {April},
    year = {2026}
}

@article{barres2025tau,
  title={tau2-Bench: Evaluating Conversational Agents in a Dual-Control Environment},
  author={Barres, Victor and Dong, Honghua and Ray, Soham and Si, Xujie and Narasimhan, Karthik},
  journal={arXiv preprint arXiv:2506.07982},
  year={2025}
}

@inproceedings{wang2022scienceworld,
  title={Scienceworld: Is your agent smarter than a 5th grader?},
  author={Wang, Ruoyao and Jansen, Peter and C{\^o}t{\'e}, Marc-Alexandre and Ammanabrolu, Prithviraj},
  booktitle={Proceedings of the 2022 Conference on Empirical Methods in Natural Language Processing},
  pages={11279--11298},
  year={2022}
}

@inproceedings{lu2025toolsandbox,
  title={Toolsandbox: A stateful, conversational, interactive evaluation benchmark for llm tool use capabilities},
  author={Lu, Jiarui and Holleis, Thomas and Zhang, Yizhe and Aumayer, Bernhard and Nan, Feng and Bai, Haoping and Ma, Shuang and Ma, Shen and Li, Mengyu and Yin, Guoli and others},
  booktitle={Findings of the Association for Computational Linguistics: NAACL 2025},
  pages={1160--1183},
  year={2025}
}

@article{liu2026planbench,
  title={PlanBench-XL: Evaluating Long-Horizon Planning of LLM Tool-Use Agents in Large-Scale Tool Ecosystems},
  author={Liu, Jiayu and Lin, Qihan and Qian, Cheng and Wang, Rui and Acikgoz, Emre Can and Yang, Xiaocheng and Liu, Jiateng and Wang, Zhenhailong and Chen, Xiusi and Ji, Heng and others},
  journal={arXiv preprint arXiv:2606.22388},
  year={2026}
}

@article{garg2026real,
  title={Real: Benchmarking autonomous agents on deterministic simulations of real websites},
  author={Garg, Div and Caples, Diego and Draguns, Andis and Ravi, Nikil and Putta, Pranav and Garg, Naman and Hebbar, Prannay and Joo, Youngchul and Gu, Jindong and London, Charles and others},
  journal={Advances in Neural Information Processing Systems},
  volume={38},
  year={2026}
}

@inproceedings{zhang2026deepplanning,
  title={Deepplanning: Benchmarking long-horizon agentic planning with verifiable constraints},
  author={Zhang, Yinger and Jiang, Shutong and Li, Renhao and Tu, Jianhong and Su, Yang and Deng, Lianghao and Guo, Xudong and Lv, Chenxu and Lin, Junyang},
  booktitle={Proceedings of the 64th Annual Meeting of the Association for Computational Linguistics (Volume 1: Long Papers)},
  pages={7377--7407},
  year={2026}
}

@article{ma2024agentboard,
  title={Agentboard: An analytical evaluation board of multi-turn llm agents},
  author={Ma, Chang and Zhang, Junlei and Zhu, Zhihao and Yang, Cheng and Yang, Yujiu and Jin, Yaohui and Lan, Zhenzhong and Kong, Lingpeng and He, Junxian},
  journal={Advances in neural information processing systems},
  volume={37},
  pages={74325--74362},
  year={2024}
}
    }    
\end{document}